# A Training-Free, Alignment-Free Approach to Corporate Intelligence: Application to SEC Filings

**Jean-François Delpech**
WebGlyphs, Inc.
Mesa, AZ 85203

High-dimensional dense text embeddings and large language models face real obstacles in financial-disclosure analysis: context-window limits, hallucination risk, high computational cost, and the arbitrary rotation of vector spaces across independently trained models. We present a training-free, alignment-free framework for corporate intelligence built on deterministic sparse seed vectors. Hashing word strings into a fixed high-dimensional basis places all documents and all temporal epochs in a common coordinate system by construction, removing any need for training or alignment. Accumulating these seed vectors across sentence contexts yields corpus-specific semantic signatures that compose linearly, supporting sub-second document comparison, issuer fingerprinting, tracking of how an issuer's vocabulary shifts between filings, and thematic sentence extraction, all on ordinary CPU hardware. Demonstrating the approach on a multi-year corpus of SEC filings (10-K, 10-Q, 8-K), we show how material corporate events, among them Boeing's 737 MAX crisis, Intel's supply-chain disruptions, and Bunge's acquisition of Viterra, emerge as distinct, interpretable semantic profiles, each traceable to the exact source sentences that produced it, with no domain-specific training and no LLM inference.

## 1. Introduction

The analysis of corporate disclosures submitted to the U.S. Securities and Exchange Commission (SEC) — annual reports, quarterly filings, and material event notices — has become a significant area of applied natural language processing. A natural approach today is to apply a large language model [1], [2], [3], which reads narrative text with an analytical depth earlier methods did not achieve. Applied directly to SEC filings, however, LLMs face several practical obstacles. A single 10-K may run to several megabytes, exceeding any practical context window, so the document must first be reduced to a high-signal subset. Confidentiality is a further concern, as routing proprietary analyses through external providers is often unacceptable; so is hallucination, since a generative model may introduce facts absent from the document, which is disqualifying in a compliance context; finally, cost may become prohibitive when processing the tens of thousands of filings published annually.

Domain-specific transformer models such as FinBERT [4] mitigate the confidentiality, cost and hallucination concerns, being deterministic classifiers that can be run locally, but they do not address the selection problem: a filing must still be reduced to the sentences worth scoring. In practice this is often done by restricting attention to a single section such as Management's Discussion and Analysis, which is limiting, as it discards the litigation notes, risk factors, and event disclosures that carry much of the material information. Earlier lexical methods such as sentiment dictionaries and keyword matching [5] are fast and transparent but share this dependence on a prior notion of what to look for, and capture little of the contextual structure of the text.

Most methods for analyzing and reducing filings represent words or sentences as dense vectors, or embeddings, learned so that terms used in similar contexts sit near one another in a high-dimensional space. Word-based embeddings such as Word2Vec assign one vector per term [6, 7], capturing its aggregate usage. In a conceptually very different approach, contextual models such as BERT [8] assign a vector to each token in context: a transformer reduces a sentence such as "Onpattro is an RNAi therapeutic." to sub-word tokens such as *on*, *##pa*, *##tt*, *##ro*, *is*, *an*, *rna*, *##i*, *therapeutic*, *.*, retrieves each token's vector, combines them as a function of position, and returns a single normalized vector for the sentence [9]. The representation of "onpattro" thus depends on its context within the sentence, but there is no notion of a corpus context: the model has no representation of how an issuer uses the word across its filings as a whole.

A further limitation afflicts embeddings obtained by training. Because they can only be defined up to a rotation, two models trained independently, even on the same data, yield spaces that are arbitrarily rotated relative to one another. Any comparison of meaning across separately trained models therefore demands an explicit, complex alignment step [10], one that introduces approximation error and grows less reliable over long time spans or between divergent corpora [11]. Such models are, moreover, neither incremental nor auditable: incorporating new documents calls for retraining, and no tractable path runs from a similarity score back to the sentences that gave rise to it.

The approach developed here sidesteps these limitations by dispensing with learned representations entirely, in favor of a fixed, shared basis. One could in principle remove the rotation problem by giving every word its own orthogonal basis vector, but that would require a dimension equal to the size of the vocabulary and is therefore intractable. Instead we exploit a property of high-dimensional geometry: while a space of dimension $D$ admits at most $D$ mutually orthogonal vectors, it admits exponentially many *quasi-orthogonal* ones, that is a set of $\exp(\mathcal{O}(\epsilon^2 D))$ randomly chosen vectors will, with high probability, have pairwise angles within $\epsilon$ of $90^\circ$ [12]. Random projection onto such a space approximately preserves distances between points [13, 14], so a modest dimension suffices to represent a large vocabulary with negligible interference between terms. We assign each word a fixed *seed vector* drawn from such a set, in dimension $D = 384$, with sixteen $+c$ and sixteen $-c$ entries with $c = \frac{1}{\sqrt{32}}$ and the rest zero [15, 16]. Because these vectors are fixed and shared, all documents and all epochs are located in the same space by construction: no training, no alignment, no rotation. Implementation details are developed in the Appendix.

Seed vectors are deterministic and semantically empty; they acquire meaning through use. For our purposes the semantic content of a word is defined entirely by its distributional neighborhood: two words are semantically similar to the degree that they occur in similar contexts, and a word's meaning in any corpus is nothing more than the statistical profile of its co-occurrences. The idea that meaning derives from use predates its computational form, as Wittgenstein [17] argued that the meaning of a word is its use in the language, but Firth gave it the formulation now standard in the field: "you shall know a word by the company it keeps" [18]. In our case, it is not a philosophical position but rather a demonstrably effective engineering choice which takes a concrete form in our linear approach: the vector of a sentence is the weighted sum of the seed vectors of its words; a frequent word such as *the* carries zero weight and a rare one such as *fluoropyrimidine* carries maximum weight. The semantic vector of a word is built by accumulating the vectors of the sentences in which it appears. This is how seed vectors give rise to *semantic vectors*, and two occurrences of the same written form are semantically distinct if and only if their distributional neighborhoods differ (see algorithm A.1 in the Appendix.) This is our operative definition of semantic distinctiveness.

This construction yields three properties that learned embeddings lack. First, a word's representation is corpus-determined: it encodes how the word is used throughout the corpus from which the vectors are built (typically an issuer's SEC filings up to a given date), not merely within one sentence. Second, because the seed vector is shared, the same word in two different corpora starts from an identical basis and diverges only through use, so its two semantic vectors are directly comparable and each faithfully reflects its own corpus environment. Third, because the vectors for words, sentences, documents, and whole corpora are all built linearly by summation in the same fixed basis, every level of the hierarchy has a compatible representation, and any two can be compared or combined directly. A further consequence follows immediately: since the bases are fixed, the semantic difference between two documents can be read off by listing the words whose vectors differ most between them (see algorithm A.6 in the Appendix.) And because the process involves only additions and multiplications, with no training or alignment, it is extremely fast: generating a word's seed vector by deterministic hashing takes a few hundred nanoseconds on a standard computer, and a full document comparison completes in well under a second.

To summarize the terminology used throughout: each word is assigned a fixed *seed vector*, deterministic and semantically empty. Accumulating the seed vectors of the sentences in which a word occurs creates a *semantic vector* which encodes the word's distributional neighborhood in a given corpus [19]. The *semantic space* of any collection of texts, a sentence, a filing, or an entire multi-year corpus, is the set of semantic vectors of its constituent words, all expressed in the common seed basis. Because every semantic vector is built by summation in that shared basis, semantic spaces compose linearly: the semantic space of a union of texts is obtained directly from those of its parts, and any linear combination of semantic spaces is itself a semantic space. It is this closure under linear combination that makes the operations of the following sections well defined: accumulating filings over time, comparing an issuer against a corpus, and projecting one space into another.

## 2. A word meaning depends on its context

Semantic vectors are therefore context-dependent: the same word carries very different meaning in the filings of a pharmaceutical company, a consumer company, or a semiconductor firm. Because vector construction relies entirely on linear additive operations over local co-occurrence windows, the resulting vector parameterizes a term's operational context rather than any abstract dictionary definition.

This holds not only for polysemous words but even for strictly monosemous terms: the representation still shifts across corpora, encoding how a particular domain uses the word within its own semantic space, as the analyses of *digital*, *cloud*, and *africa* below illustrate. The similarity between two such vectors $\vec{u}$ and $\vec{v}$ is the cosine similarity $\sigma_{uv} = \vec{u} \cdot \vec{v} / \|\vec{u}\| \, \|\vec{v}\|$; where a metric is more convenient we use the associated distance $\mathcal{D}_{uv} = \sqrt{2(1 - \sigma_{uv})}$. The linearity of the construction and the Euclidean nature of this distance carry several useful properties for our purposes, developed in the Appendix.

### *2.1. Neighbors of 'digital'*

**Table 1 - Neighbors of *digital* in different corpora**

|  |  |
|---|---|
| **Alnylam Pharmaceuticals** | *Healthcare/Biotechnology* |
| | digital, counties, examining, economy, considering, oecd, taxing, individual, corporations, multinational, co_operation, allocated, recently, erosion, taxation |
| **AeroVironment** | *Industrials/Aerospace & Defense* |
| | digital, mechanical, water, zoom, pan, tilt, infrared, bungee, landing, launch, vertical, lightweight, aerostructures, hand, autonomous |
| **Mercury Systems** | *Industrials/Aerospace & Defense* |
| | digital, examples, input_output, memory, circuits, mmics, converters, amplifiers, monolithic, switches, analog, chips, limiters, equalizers, oscillators |
| **Advanced Micro Devices** | *Technology/Semiconductors* |
| | digital, circuits, buffer, analog, signal, transmitters, serializing, paths, transceivers, switched, convertors, conditioners, switching, comparators, adi |
| **Bunge Global SA** | *Consumer Defensive/Farm Products* |
| | digital, misuse, interception, manually, coca_cola, trustee, vision, commerce, kellogg, spin, corteva, dowdupont, record_keeping, authenticates, signatures |

Although the polysemous word *digital* appears in all of these filings, its neighbors vary dramatically with each company's business model, product taxonomy, and regulatory context. For Alnylam, *digital* is anchored in policy language addressing the OECD's Digital Economy and Base Erosion and Profit Shifting (BEPS) tax initiatives; it appears almost entirely in disclosures about international tax reform on multinational revenues. AeroVironment, which manufactures small tactical drones, uses it for physical payload components: digital electro-optical and infrared pan-tilt-zoom camera gimbals and flight-control hardware. For Mercury Systems it belongs to hardware-engineering terms, digital-to-analog converters, MMICs, and high-speed digital circuit architectures. AMD uses it similarly, in the context of chip-level mixed-signal design, high-speed transceivers and circuits; ADI (Analog Devices, Inc.) was a key intellectual property rival engaging in high-stakes IPR challenges at the time AMD acquired Xilinx. Bunge, a global agricultural commodity trader, uses *digital* in yet another sense entirely: e-commerce platforms, digital signatures for grain contracts, system-downtime risk. The appearance of *coca_cola* and *kellogg* among Bunge's neighbors for *digital* has a more prosaic origin: they occur in the biography of a board nominee whose prior roles at Kellogg, Coca-Cola, and Tyson included responsibility for digital media and e-commerce. The distributional method faithfully captures this, since *digital* is genuinely used near those company names in Bunge's filings, but the association is biographical rather than operational.

The last point shows the method's fidelity and its limits in one example: the vector correctly reflects the text, but the analyst must still interpret *why* two terms are neighbors. A biographical and an operational co-occurrence may look identical and they can be distinguished only by reading and understanding the sentence. This limitation, however, is not particular to our approach, it is shared by any method that remains faithful to the source text, including a large language model, which would encounter the same biographical passage and form the same association. An LLM might, if suitably prompted, recognize that the names come from a director's résumé rather than from Bunge's operations; but no prompt guarantees this, and the plausible-but-false alternative of an invented commercial link between a commodity trader and a beverage company is exactly the kind of association such models tend to supply. Our method does neither: it reports the co-occurrence traceably, never invents an association absent from the filing, and leaves to the analyst the interpretive step, as it should.

### *2.2. Neighbors of cloud*

The neighbors of *cloud* (next page) do not separate along the familiar English polysemy of weather versus computing. Instead they expose a structural division between accounting-standard disclosure and operational product infrastructure. In the non-semiconductor filings (Alnylam, AeroVironment, Mercury), *cloud* almost never denotes a product; it appears as boilerplate surrounding FASB Accounting Standards Updates on the treatment of cloud-computing arrangements, and its neighbors, *fasb*, *asu*, *implementation*, *capitalize*, *retrospective*, map it to a balance-sheet and disclosure context. For AMD, a semiconductor and hardware company, *cloud* shifts from an accounting line item to a primary revenue driver: its neighbors, *azure*, *aws*, *oracle*, *google*, *tencent*, *hyperscale*, *workloads*, reconstruct the actual hyperscaler ecosystem in which the company operates.

Bunge is the informative negative case: its filings do not contain the word *cloud* in the period covered, and the method returns an empty neighbor set rather than a list of weakly related terms. This is a property worth emphasizing: where a term is genuinely absent from a corpus, the linear construction yields a null result, avoiding the artificial similarity inflation that dense pre-trained embeddings produce when forced to place every term somewhere in an occupied space.

**Table 2 - Neighbors of *cloud* in different corpora**

|  |  |
|---|---|
| **Alnylam Pharmaceuticals** | *Healthcare/Biotechnology* |
| | cloud, computing, heightens, hosting, fasb, interaction, clarify, implementation, debt_debt, shortening, callable, simplifying, outlines, disaggregation, arrangements |
| **AeroVironment** | *Industrials/Aerospace & Defense* |
| | cloud, computing, implementation, stock_based, arrangement, depreciation, amortization, activity, non_operating, expenses, compensation, accounting, ebitda, add, non_cash |
| **Mercury Systems** | *Industrials/Aerospace & Defense* |
| | cloud, clarified, asu, early, adoption, portray, implementation, capitalize, modifying, enact, classifying, retrospective, incurred, clarify, sunset |
| **Advanced Micro Devices** | *Technology/Semiconductors* |
| | cloud, oracle, azure, google, aws, amazon, epyc, gen, infrastructure, solutions, workloads, siemens, instinct, tencent, creo |
| **Bunge Global SA** | *Consumer Defensive/Farm Products* |
| | *Null subspace* |

### *2.3. Neighbors of africa*

A word need not be polysemous for the linear vector method to discriminate. Even a term with a single, unambiguous dictionary sense will produce divergent vectors across industries when the surrounding vocabulary differs. What the vector encodes is not that fixed sense but the word's usage in each corpus. The list on next page gives, in decreasing order of semantic similarity, the neighbors of *africa* in different corpora.

While *africa* denotes the same continent in every filing, its nearest neighbors are very different depending on industry lines, business units, and reporting habits. Alnylam maps Africa as an administrative territory within its CEMEA commercial region (Central and Eastern Europe, Middle East, Africa), where it appears among other distribution geographies, *italy*, *belgium*, *france*, *turkey*. AeroVironment views it through defense-market dynamics, specifically the MENA demand landscape in a post-Ukraine spending environment, with neighbors drawn from investor-presentation vocabulary (*cagr*, *slide*, *conference*). Mercury Systems projects the region onto its defense hardware, signal processing, tactical edge computing, airspace-domain monitoring, under the EMEA grouping. Bunge ties Africa to commercial client management and corporate strategy, its neighbors mixing regional terms with the names of executives and advisers involved in operational reviews.

| Table 3 - Neighbors of *africa* in different corpora | |
|---|---|
| **Alnylam Pharmaceuticals** | *Healthcare/Biotechnology* |
| | africa, italy, belgium, cemea, turkey, sweden, atu, france, germany, middle, europe, ast, canada, portugal, spain |
| **AeroVironment** | *Industrials/Aerospace & Defense* |
| | mena, africa, post_ukraine, cagr, aerovironment, norell, pacific, slide, europe, reiterating, conference, start_ups, monday, inc_june, inc_september |
| **Mercury Systems** | *Industrials/Aerospace & Defense* |
| | africa, emea, processing, bringing, edge, leaning, solve, value_add, comprises, signal, ranging, networking, soc, domains_air, middle |
| **Advanced Micro Devices** | *Technology/Semiconductors* |
| | *Null subspace* |
| **Bunge Global SA** | *Consumer Defensive/Farm Products* |
| | africa, middle, isle, man, weymouth, darren, abdellah, agouzoul, asia_pacific, south, deutsche, michael, client, head, printed |

AMD is the null case: its filings carry no significant co-occurrence signal for *africa*, and the method returns an empty neighbor set rather than a forced list of weak associations.

## 3. Semantic profiles and fingerprints

The additive nature of semantic spaces has a direct operational consequence for corporate profiling. Because every document in the corpus, whether a single filing or the aggregate of all filings across all issuers, is represented as a vector in the same fixed basis, the distinctive vocabulary of any one issuer can be isolated by elementary arithmetic operations.

For example, we construct an issuer-specific distinctive-term profile (or fingerprint) by contrasting the language of all of an issuer's filings over a selected period against a reference SEC corpus (see algorithm A.5 in the Appendix). In a small fraction of a second this yields a ranked list of the terms most characteristic of that issuer relative to the corpus. These fingerprints, here covering June 2019 to July 2025, reveal issuer-specific information that a generic financial summary often obscures: product portfolios and technology architectures, acquisitions and collaboration partners, facilities, executive names, and legal entities.

AeroVironment is a useful example for a less widely followed issuer. Terms such as *bluehalo*, *telerob*, *hapsmobile*, and *switchblade* point to acquisitions, autonomy and unmanned-system capabilities, strategic relationships, and product lines. This intelligence would otherwise require time-consuming manual review to reconstruct, whereas the fingerprint updates almost instantaneously as a new filing enters the corpus. AeroVironment's acquisition of BlueHalo, completed in May 2025 [20], shows how a major corporate event appears directly in the profile.

| Table 4 - Examples of fingerprints (July 2025) | |
|---|---|
| **Alnylam Pharmaceuticals** | *Healthcare/Biotechnology* |
| | rnai^65.1, onpattro^60.7, givlaari^52.6, oxlumo^35.2, vutrisiran^24.4, amvuttra^21.9, zilebesiran^19.8, inclisiran^17.8, hattr^17.0, fitusiran^16.2, patisiran^15.6, sirnas^13.0, cemdisiran^12.4, mdco^8.8, lumasiran^6.1, attr^5.9, revusiran^5.0 |
| **AeroVironment** | *Industrials/Aerospace & Defense* |
| | muas^24.9, bluehalo^21.5, telerob^17.2, hapsmobile^15.1, webasto^11.2, haps^10.3, ugv^8.7, suas^7.2, uxs^6.1, simi^5.3, tomahawk^5.2, nawabi^4.1, switchblade^3.9, maccready^3.3, avav^3.3, wahid^3.0, ums^3.0 |
| **Mercury Systems** | *Industrials/Aerospace & Defense* |
| | andover^9.0, ballhaus^5.6, minuteman^4.6, aslett^4.1, pentek^4.0, first_lien^3.8, ruppert^3.5, acton^3.1, avalex^2.8, expense^2.7, jana^2.7, proprietary_no^2.5, farnsworth^2.5, starboard^2.4, recently^2.2, geco^1.8, syntonic^1.6 |
| **Advanced Micro Devices** | *Technology/Semiconductors* |
| | radeon^13.0, atmp^8.5, pensando^4.6, thatic^4.5, zen^3.9, rdna^3.8, threadripper^3.2, instinct^3.0, athlon^2.1, wessels^1.9, tongfu^1.8, rocm^1.6, soms^1.4, collabo^1.4, higon^1.4, sanmina^1.2, add_in_board^1.2 |
| **Bunge Global SA** | *Consumer Defensive/Farm Products* |
| | danube^20.2, bafc^8.7, manor^4.7, cooperatieve^4.6, timberlake^4.5, rabobank^4.4, bfe^4.3, loders^3.8, us_active^3.8, chesterfield^3.4, utilisation^3.1, cobank^1.8, bye_laws^1.5, sugarcane^1.5, ffa^1.4, croklaan^1.3, heckman^1.2 |

Alnylam and AMD illustrate how sharply the method separates issuers by their actual product portfolios. Alnylam's fingerprint is almost entirely its RNAi franchise, *rnai*, *onpattro*, *givlaari*, *oxlumo*, *vutrisiran*, *amvuttra*, a compact map of both marketed medicines and development-stage assets that no competitor would reproduce; Onpattro, the first FDA-approved RNA-interference therapy [21], is historically meaningful rather than merely lexical. AMD's is organized around product families, computing architectures, and acquired or partnered capabilities, *radeon*, *pensando*, *zen*, *rdna*, *threadripper*, *instinct*, tracing both its organic roadmap and its acquisition history.

Mercury Systems shows a different kind of signature. Its fingerprint combines a location and facility identifier (*andover*, *minuteman*), an acquisition (*pentek*), financing language (*first_lien*), and executive names (*ballhaus*, *ruppert*, *aslett*) — a compact record of corporate history and organizational context, demonstrating that a useful fingerprint need not be limited to consumer-facing product names.

In a production environment these fingerprints are not static lists but entry points. An analyst calling up a company's profile can select any term, or group of terms, and retrieve the verbatim SEC passages in which they occur, moving directly from the compact fingerprint to the exact sentences of the source filings. While analyzing Bunge's, entering *ukr** in the search field retrieves every sentence containing a word that begins with those letters, *ukraine*, *ukrainian*, *ukrainian_based*, *ukrainian_exposed*, *ukraine_russia*, a prefix search of the semantic space that no pooled representa-tion could support, since it depends on the individual words still being present in the index along with their localization down to the filing's level.

This is a direct benefit of retaining the identity of every word rather than collapsing the text into a pooled vector: each term resolves to the specific filings, sections, and sentences that produced it. The fingerprint thus serves as a navigable index into the underlying disclosure database, letting an analyst confirm in seconds what a term means and in what context the issuer used it, rather than taking the profile on trust. And because the distinctive terms are precisely the ones that name a company's specific products, entities, and events, they double as high-quality external search queries: dropped into a web search, any of them (such as 'amd thatic') pulls in the outside context, news, analyst commentary, regulatory action, that the filings themselves do not contain.

Note that intuitively, the semantic dispersion of fingerprint words carries information about the issuer, and may relate in part to firm complexity in the sense studied by Loughran and McDonald [22]. We touch on this in Appendix A.2.4.

## 4. Time Evolution of SEC Filings

In addition to the static profiles discussed in the previous section, our method can equally well generate dynamic profiles every time an issuer publishes a filing: to analyze a filing published at date $t$, we simply compare the two cumulative semantic spaces at dates $t$ and $t-1$. This tells us how much each word's semantic vector grew between the two dates, and we keep those whose contribution increased significantly, since they are the words the latest filing reinforced or introduced, and they obviously characterizes what is new or intensifying in the disclosure (see algorithm A.6 in the Appendix.)

Having selected this set of distinctive words, we compute the pairwise similarities of their semantic vectors and use the resulting similarity matrix to group them by agglomerative hierarchical clustering with linkage average. Each cluster member gathers words that occur in similar contexts, so it corresponds to a coherent theme within the filing, and we retrieve for each, by exact term matching, the sentences in which the selected words appear. This results in a compact, thematically organized set of fully tagged and ordered verbatim sentences illustrating the filing's themes in decreasing order of distinctiveness; in other words, we cluster first the distinctive *words* and then retrieve corresponding sentences afterward, so each cluster and cluster element is already labelled by the terms that produced it.

This clustering approach differs from first clustering embedded sentence vectors, whether from BGE or from our own semantic vectors used at the sentence level. In that alternative, sentences are grouped by mutual similarity, which works well but is more generic and yields clusters of sentences without a set of characteristic words to label them: the grouping is available, but not the vocabulary defining each theme. Our construction retains the advantage of the fixed shared basis: the representations are comparable across filings and corpora without alignment, and every cluster is traceable to the source words.

In all the following tables, clusters are generated from words; for reasons of space, only the most significant words, cluster elements, and sentences are shown. In a production setting these parameters are adjustable and most elements are clickable, so that clusters can be shown at much greater depth, giving fuller coverage of each theme while preserving the same traceability from cluster to term to source sentence.

The tables are, in fact, only summary views of an underlying database: for each filing we retain the full set of clusters with all their relevant sentences, keywords, weights, and named entities, together with the identifiers needed to link each element back to its source. Any of these can be called up on demand, and the same stored representation supports cross-issuer comparisons and the other analyses presented below.

### *4.1. Mercury Systems 2021-07-02 10-K*

| **Table 5 - Distinctive words and extracts from Mercury Systems (*MRCY*) 10-K_20210702** |
|---|
| Cluster #1 - Top words: pentek, impact, collectively, radio, management, manufacturer, osha, upper, saddle, river, report, etc. |
| ("Athena"), Delta Microwave, LLC ("Delta"), Syntonic Microwave LLC ("Syntonic"), and Pentek Technologies, LLC and Pentek Systems, Inc. (collectively, "Pentek"); mission computing, safety-critical avionics and platform management, and large area display technology with the CES Creative Electronic Systems, S.A. |
| The acquisitions of the Carve-Out Business, Delta, Syntonic and Pentek further improved our ability to compete successfully in these market segments by allowing us to offer an even more comprehensive set of closely related capabilities. |
| Over the lifetime of a defense program, the award of many different individual contracts and subcontracts may impact our products' requirements. |
| Based in Upper Saddle River, New Jersey, Pentek is a leading designer and manufacturer of ruggedized, high-performance, commercial off-the-shelf ("COTS") software-defined radio and data acquisition boards, recording systems and subsystems for high-end commercial and defense applications. |
| The impact to income taxes includes the impact to the effective tax rate, current tax provision and deferred tax provision. |
| The following table presents the net purchase price and the fair values of the assets and liabilities of Pentek on a preliminary basis: Consideration transferred Cash paid at closing $ 65,668 Less cash acquired ( 746) Net purchase price $ 64,922 Estimated fair value of tangible assets acquired and liabilities assumed Cash 746 Accounts receivable 1,303 Inventory 6,522 Fixed assets 152 Other current and non-current assets 2,864 Accounts payable ( 1,016) Accrued expenses ( 520) Other current and non-current liabilities (3,718) Estimated fair value of net tangible assets acquired 6,333 Estimated fair value of identifiable intangible assets 24,110 Estimated goodwill 35,225 Estimated fair value of net assets acquired 65,668 Less cash acquired ( 746) Net purchase price $ 64,922 The amounts above represent the preliminary fair value estimates as of July 2, 2021 and are subject to subsequent adjustment as the Company obtains additional information during the measurement period and finalizes its fair value estimates. |
| Cluster #2 - Top words: andover, subsystems, massachusetts, apps, kong, hong, aerospace, trusted, incorporation, environments, processing, etc. |
| Our Andover, Massachusetts and Hudson, New Hampshire facilities design and assemble our processing products and are AS9100 quality systems-certified facilities. |
| Our Andover, Massachusetts facility is also a DMEA-certified trusted design facility and is primarily focused on advanced security features for the processing product line. |
| For example, If the US government continues to expand the scope of regulations intended to address civil-military fusion in China, certain commercial technologies which have historically been exportable to Hong Kong and China may be prohibited. |
| We also are subject to the Massachusetts General Laws which, subject to certain exceptions, prohibit a Massachusetts corporation from engaging in a broad range of business combinations with any "interested shareholder" for a period of three years following the date that such shareholder becomes an interested shareholder. |
| Cluster #3 - Top words: perry, annual, aslett, suntron, brian, exhibit, section, reference, incorporated, item, amended, etc. |
| Certain factors that might cause such a difference are discussed in this annual report on Form 10-K, including in the section entitled "Risk Factors." |
| Mr. Aslett has also held positions at GEC Plessey Telecommunications, as well as other telecommunications-related technology firms. |
| Prior to joining Mercury, Mr. Perry was the General Manager for Suntron Corporation's Northeast Express and served in various roles with Lockheed Martin and General Electric Aircraft Engines. |
| Cluster #4 - Top words: mazzola, document, ruppert, interactive, embedded, select, matters, title, par, code, principal, etc. |
| *etc.* |

In Table 5 the first cluster assembles the vocabulary of a specific acquisition, gathering the target's name, location, and product line (*pentek*, *upper*, *saddle*, *river*, *radio*, *manufacturer*, *osha*), while the second gathers the company's own corporate and operational identity (*andover*, *massachusetts*, *aerospace*, *headquartered*, *environments*).

This set, extracted in a fraction of a second from a 2.7 MB 10-K filing, shows how linear vectors enable high-discrimination word scores and purely programmatic extraction of a document's core theme. Every top-ranked term corresponds directly to Pentek's company profile, location, or technical footprint, and the extracted sentences together cover the three elements an analyst needs from an M&A disclosure: the target's identity and capabilities, the strategic rationale, and the transaction mechanics.

Note the unexpected prominence of *saddle*, which here has nothing to do with its ordinary meaning: it appears because Pentek is headquartered in Upper Saddle River, New Jersey, and the term occurs almost exclusively in Mercury's filings. It is a striking illustration of the method's central mechanism, a token becoming distinctive purely through corpus contrast, regardless of its dictionary sense, and one that no predefined keyword list or industry taxonomy would have included.

### *4.2. INTEL 10-Q 2020-03-28*

This fingerprint captures the vocabulary of major COVID disclosures, with distinctive words such a *pandemic*, *suppliers*, *coronavirus*, terms that didn't appear in earlier SEC filings. The containment measure vocabulary is listed: travel bans, quarantines, shelter-in-place, social distancing, shutdowns.

**Table 6 - Distinctive words and extracts from Intel Corporation (*INTC*) 10-Q_20200328**

| Cluster #1 - Top words: covid, pandemic, adversely, affect, spread, condition, restrictions, coronavirus, cause, suppliers, travel, etc. |
| --- |
| As we navigate through the effects of the COVID-19 pandemic we are working to ensure compliance with orders and restrictions imposed by government authorities. |
| The novel strain of the coronavirus identified in China in late 2019 (COVID-19) has globally spread throughout other areas such as Asia, Europe, the Middle East, and North America and has resulted in authorities imposing, and businesses and individuals implementing , numerous unprecedented measures to try to contain the virus, such as travel bans and restrictions, quarantines, shelter - in - place/stay-at-home and social distancing orders, and shutdowns. |
| It is likely that the current outbreak and continued spread of COVID-19 will cause an economic slowdown, and it is possible that it could cause a global recession. |
| The spread of COVID-19 has caused us to modify our business practices (including employee travel, employee work locations, cancellation of physical participation in meetings, events and conferences, and social distancing measures), and we may take further actions as may be required by government authorities or that we determine are in the best interests of our employees, customers, partners, vendors, and suppliers. |
| Our operations and business, and those of our customers and suppliers, can be disrupted by natural disasters; industrial accidents; public health issues (i ncluding the COVID-19 pandemic discussed further in the risk factor "The COVID-19 pandemic could materially adversely affect our financial condition and results of operations," above); cybersecurity incidents; interruptions of service from utilities, transportation, telecommunications, or IT systems providers; manufacturing equipment failures; or other catastrophic events. |
| Evaluation of Disclosure Controls and Procedures Due to the COVID-19 pandemic, a significant portion of our employees are now working from home, while also under shelter-in-place orders or other restrictions. |
| Cluster #2 - Top words: california, employer, mission, college, clara, santa, principal, former, code, offices, charter, etc. |
| *etc.* |

The distinctive words mostly belong to three categories, fab-specific risk related to employees contracting the virus and disrupting manufacturing, key personnel unavailability risk, and generic economic recovery uncertainty.

### *4.3. Bunge20230612*

| **Table 7 - Distinctive words and extracts from Bunge Global SA (*BG*) 8-K_20230612** |
|---|
| Cluster #1 - Top words: shall, danube, section, cppib, except, viterra, affiliates, cfius, subsidiaries, examinership, ifrs, etc. |
| Ocorian Limited, a company incorporated in Jersey in its capacity as trustee of the Viterra Employee Benefit Trust, a trust for the benefit of certain current and former service providers of Viterra (collectively with Glencore, CPPIB and BCI, the " Sellers " and each individually, a " Seller "). |
| Danube Subsidiaries " means the Subsidiaries of Danube (which, for the avoidance of doubt, shall include Gavilon Agriculture Investment, Inc. and any of its Subsidiaries). " |
| Recovery Costs " means any Losses, punitive damages, or exemplary damages, of or incurred by Amazon, Danube, any Amazon Subsidiary, or any Danube Subsidiary, in connection with or otherwise related to any Matters (as defined in Section 6.7 of the Danube Disclosure Letter) or the CPPIB Subscription Letter, or Amazon's and Danube's performance of their obligations under Section 6.7 . " |
| Cluster #2 - Top words: communications, beneficial, gotshal, weil, e_c, soliciting, instruction, subsequent, equivalents, reverse, depreciation, etc. |
| Amazon IT Systems " means computers, Software, hardware, servers, networks, routers, hubs, switches, data communications lines, data storage devices, and other information technology equipment owned, leased or licensed by Amazon or any of the Amazon Subsidiaries and used in connection with their respective businesses. " |
| Articles of Association " means the Company's Articles of Association, dated [ ] (as may be amended from time to time). " beneficial ownership " and related terms such as "beneficially owned" or "beneficial owner" have the meanings given such terms in Rule 13d-3 under the Exchange Act and a Person's beneficial ownership of Capital Stock shall be calculated in accordance with the provisions of such rule. " |
| Cluster #3 - Top words: florissant, lisa, chesterfield, parkway, manor, timberlake, irs, offices, epidemics, temporary, pandemics, etc. |
| Except as would not reasonably be expected to, individually or in the aggregate, be materially adverse to Danube and its Subsidiaries, taken as a whole, Danube and the Danube Subsidiaries are, and in the past three (3) years have been, in material compliance with all Laws relating to labor, employment and/or employment practices, including all applicable Laws relating to hiring, background checks, sexual harassment training, wages, hours, overtime, pay equity, immigration, employment eligibility verification, collective bargaining, labor relations, employment discrimination, harassment, retaliation, privacy, whistleblowing, disability rights and benefits, sick time, leaves of absences, safety and health, COVID-19, terminations, plant closures and mass layoffs, workers' compensation, and classification of exempt employees, temporary employees, outsourced employees, independent contractors and other non-employee workers. |
| All notices and other communications hereunder shall be in writing and shall be deemed given if delivered personally, telecopied or sent by email transmission (so long as an error message is not generated in reply thereto) or sent by registered or certified mail, postage or by prepaid overnight courier, to the Parties at the following addresses (or at such other address for a Party as shall be specified by like notice): if to the Company, to: Bunge Global SA Route de Florissant 13 Geneva 1206, Switzerland Attention: Joseph Podwika Lisa Ware-Alexander Email: [***] with a copy to (which shall not constitute notice): Latham & Watkins LLP 1271 Avenue of the Americas New York, NY 10020 Attention: Charles K. Ruck Max Schleusener Email: charles.ruck@lw.com max.schleusener@lw.com If to the Shareholder, to: Danelo Limited Baarermattstrasse 3, PO Box 6341 Baar, Switzerland Attention: Shaun Teichner John Burton Email: [***] 25 with a copy to (which shall not constitute notice): Weil, Gotshal & Manges LLP 767 Fifth Avenue New York, NY 10153 Attention: Michael J. Aiello Douglas P. Warner David Avery-Gee Email: michael.aiello@weil.com doug.warner@weil.com david.avery-gee@weil.com Section 10.5 Interpretation . |
| Cluster #4 - Top words: nonactions, cautionary, embedded, cover, interactive, differ, unavailable, inability, negatively |
| *etc.* |

The extraction of Bunge's June 12, 2023 8-K captures the Viterra merger agreement. Two of the most prominent terms, *danube* and *amazon*, require a word of explanation, since neither means what it appears to. They are code names used throughout the agreement in place of the actual parties: *danube* designates Bunge and its subsidiaries (Gavilon among them), and *amazon* designates the Viterra side. Merger agreements routinely substitute such placeholders during negotiation to preserve confidentiality, and the defined-term language is carried into the executed document. The method identifies these terms because they are distinctive and pervasive in the filing, but it cannot know that *danube* is Bunge or *amazon* is Viterra; that identification requires reading the definitions, which the extracted sentences supply ("Danube Subsidiaries means the Subsidiaries of Danube, which shall include Gavilon..."). This is a clean illustration of the division of labor the method assumes: it locates and clusters the distinctive vocabulary and returns the exact defining sentences, leaving the interpretation of what a code name denotes to the reader or to a downstream model working from those sentences.

*4.4. Boeing*

| **Table 8 - Distinctive words and extracts from The Boeing Company (*BA*) 8-K_20210106** |
|---|
| Cluster #1 - Top words: max, faa, airplane, employees, employee, aircraft, calhoun, airline, training, internal, determination, etc. |
| Through this deception, the Company interfered with the FAA AEG's lawful function to evaluate MCAS and to include information about MCAS in the 737 MAX FSB Report, and fraudulently obtained from the FAA AEG a differences-training determination for the 737 MAX that was based on incomplete and inaccurate information about MCAS; b. |
| The Fraud Section determined that an independent compliance monitor was unnecessary based on the following factors, among others: (i) the misconduct was neither pervasive across the organization, nor undertaken by a large number of employees, nor facilitated by senior management; (ii) although two of the Company's 737 MAX Flight Technical Pilots deceived the FAA AEG about MCAS by way of misleading statements, half-truths, and omissions, others in the Company disclosed MCAS's expanded operational scope to different FAA personnel who were responsible for determining whether the 737 MAX met U.S. federal airworthiness standards; (iii) the state of the Company's remedial improvements to its compliance program and internal controls; and (iv) the Company's agreement to meet with and report to the Fraud Section as set forth in Attachment D to this Agreement (Enhanced Reporting Requirements); i. |
| As part of this evaluation and approval process, the FAA had to make two distinct determinations: (i) whether the airplane met U.S. federal airworthiness standards; and (ii) what minimum level of pilot training would be required for a pilot to fly the airplane for a U.S.-based airline. |
| From at least in and around November 2016 through at least in and around December 2018, in the Northern District of Texas and elsewhere, Boeing, through Boeing Employee-1 and Boeing Employee-2, knowingly, and with intent to defraud, conspired to defraud the FAA AEG. |
| Nevertheless, if Boeing did not fix the 737 MAX's pitch-up characteristic in high-speed, wind-up turns, the FAA could determine that the 737 MAX did not meet U.S. federal airworthiness standards. |
| Boeing Employee-1 and Boeing Employee-2 similarly understood that it was their responsibility to update the FAA 10 AEG about any relevant changes to the 737 MAX's flight controls-such as MCAS's expanded operational scope. |
| Cluster #2 - Top words: deferred, prosecution, section, grounding, airplanes, criminal, procedures, relating, certification, subsidiaries, conduct, etc. |
| The Fraud Section, however, may use any information related to the conduct described in the attached Statement of Facts against the Company: (a) in a prosecution for perjury or obstruction of justice; (b) in a prosecution for making a false statement; (c) in a prosecution or other proceeding relating to any crime of violence; or (d) in a prosecution or other proceeding relating to a violation of any provision of Title 26 of the United States Code. a. |
| The Fraud Section further agrees that if the Company fully complies with all of its obligations under this Agreement, the Fraud Section will not continue the criminal prosecution against the Company described in Paragraph 1 and, at the conclusion of the Term, this Agreement shall expire. |
| The Company (a) acknowledges the filing of the one-count Information (as such term is described/defined in the Agreement) charging the Company with one count of Conspiracy to Defraud the United States, in violation of Title 18, United States Code, Section 371; (b) waives indictment on such charges and enters into a deferred prosecution agreement with the Fraud Section; and (c) agrees to pay a Total U.S. Criminal Monetary Amount of $2,513,600,000 under the Agreement with respect to the conduct described in the Information; 2. |
| Cluster #3 - Top words: airlines, ethiopian, lion, exhibit, beneficiaries, heirs, accidents, amounts, page, interactive, document, etc. |
| The DPA contemplates that the Company will: (1) make payments totaling $2.51 billion, which consist of (a) a $243.6 million criminal monetary penalty; (b) $500 million in additional compensation to the heirs and/or beneficiaries of those who died in the Lion Air Flight 610 and Ethiopian Airlines Flight 302 accidents; and (c) $1.77 billion to the Company's airline customers for harm incurred as a result of the grounding of the 737 MAX, offset in part by payments already made and the remainder satisfied through payments to be made prior to the termination of the DPA; (2) review its compliance program for implementation of continuous improvement efforts; and (3) implement enhanced compliance reporting and internal controls mechanisms. |
| The Boeing Company ("Boeing") was a U.S.-based multinational corporation that designed, manufactured, and sold commercial airplanes to airlines worldwide. |
| Boeing's airline customers included major U.S.-based airlines headquartered in the Northern District of Texas and elsewhere. |
| On March 10, 2019, Ethiopian Airlines Flight 302, a Boeing 737 MAX, crashed shortly after takeoff near Ejere, Ethiopia. |
| Cluster #4 - Top words: chicago, plaza, riverside, seattle, principal, delaware, gerry, offices, specified, incorporation, irs, etc. |
| *etc.* |

The extraction of Boeing's January 6, 2021 8-K illustrates the method on a filing of unusual legal density. The 8-K and its attachments, totalling 0.318 MB, were reduced to labelled clusters in 0.142 seconds, and the clusters cleanly separate the filing's distinct subjects.

Cluster element #1 collects the conduct at issue: the terms *max*, *faa*, *airplane*, *training*, and *determination* gather the sentences describing the FAA Aircraft Evaluation Group's airworthiness and pilot-training determinations for the 737 MAX, and the alleged misrepresentations concerning the MCAS system. Cluster element #2 collects the legal resolution, with *deferred*, *prosecution*, *criminal*, *certification*, and *conduct* gathering the deferred-prosecution-agreement mechanics, including the single count of conspiracy to defraud the United States and the conditions under which the charge would be dismissed. Cluster element #3 isolates the financial settlement, its terms *airlines*, *ethiopian*, *lion*, *beneficiaries*, *heirs*, and *accidents* gathering the sentences that itemize the $2.51 billion total: airline-customer compensation, the crash-victim beneficiary fund for Lion Air Flight 610 and Ethiopian Airlines Flight 302, and the criminal penalty. Cluster element #4 collects administrative and signature-page material, with *chicago*, *plaza*, *riverside*, *delaware*, and *irs* identifying the corporate address, state of incorporation, and routine exhibit metadata.

Two observations bear on the method rather than on Boeing.

First, the clusters correspond to genuinely different registers within a single filing: the operational conduct, the prosecutorial framework, the monetary settlement, and the administrative boilerplate. The word-first clustering recovers each with a coherent, self-labelling vocabulary and its supporting verbatim passages. Because a high-distinctiveness term usually belongs to a single theme, whereas a sentence typically mixes several, clustering the words rather than the sentences sharpens the separation and gives each cluster an interpretable label by construction.

The gain runs in both directions. Consider first two bags of distinctive words, *ruppert select matters title par code principal offices intend probable expressions*, from the administrative cluster of the Mercury example, and *deferred prosecution section grounding airplanes criminal procedures relating certification subsidiaries conduct*, from the legal cluster of the Boeing example. These are semantically unrelated, one administrative boilerplate, the other criminal-settlement vocabulary, yet their pooled BGE D-384 sentence embeddings have similarity 0.67, whereas in our semantic space the similarity is 0.08. BGE is transformer-based and order-aware, so it does not treat its input as a true bag of words; token order and grammatical context condition the resulting embedding, and a nonzero similarity between two word lists is not in itself surprising. What cannot be determined is why the value is as high as 0.67: the score follows from the model's training, with no way to trace it to any feature of the text that would account for it. Our word-level similarity, by contrast, is a transparent function of the vocabulary, and here it correctly reports that the two lists have little in common.

The converse case is equally telling. Take two word lists drawn from the same Boeing filing, *max faa airplane employees employee aircraft calhoun airline* and *training internal determination connection evaluation controls obligations*. They share no word at all, so a boolean or keyword comparison would score them at zero; yet our method assigns them a similarity of 0.20, correctly recognizing that they occupy related regions of *this particular* filing's semantic space. The representation captures a relatedness that lies in shared context rather than shared tokens, which neither exact matching nor a pooled embedding recovers in an interpretable way.

Second, the extraction is fully attributed: each cluster resolves to specific sentences in the source document, so the output can serve directly as a grounded basis for an analyst or as constrained input to a language model required to quote the retrieved passages rather than summarize from memory.

The consistency of these extractions, across filings as structurally different as Mercury's, Intel's, and Boeing's, shows that the method does not rely on a fixed document format.

As noted earlier, these clusters are generated from a searchable database in which words, groups of words, and whole sentences are all represented in the single shared space, and can therefore serve interchangeably as queries. The next section develops this, showing how a query, whether a term, a set of terms, or a sentence, retrieves its neighbors and the passages that support them.

## 5. Semantic vector querying

Since every entity, word, phrase, query, sentence, or document, is represented by a semantic vector in the same shared space, any search reduces to finding the neighbors of a query vector in a database of semantic spaces.

For example, a search for the neighbors of *onpattro* in Alnylam's semantic space returns *onpattro*, *achieved*, *revenues*, *excellent*, *territories*, *updates*, *launches*, *grew*, *fourth*, *anticipated*, and *began*, a list that reads like a victory communiqué. This seems fitting, given Onpattro's role in Alnylam's history: it was the company's first product to receive marketing approval and the first FDA-approved RNAi [21] therapeutic. The distributional neighbors thus capture not merely what Onpattro. is, but the commercial and celebratory register in which the company discusses it.

This is borne out further when the returned terms are used as a query restricted to 2021 Alnylam filings, which returns:

- Fourth Quarter 2020 and Recent Significant Corporate Highlights Commercial Performance ONPATTRO Achieved global net product revenues for the fourth quarter and full year 2020 of $90 million and $306 million, respectively, representing 9.5% quarterly growth compared to Q3 - including 10.4% growth in the U.S. market segment driven by new patient demand - and over 80% annual growth from full year 2019.
- In addition, Alnylam today reported preliminary fourth quarter and full year 2020 global net product revenues for ONPATTRO and GIVLAARI and provided additional updates on the Company s commercial launches, including initial OXLUMO demand.
- Third Quarter 2021 and Recent Significant Corporate Highlights Commercial Performance ONPATTRO (patisiran) Achieved global net product revenues for the third quarter of 2021 of $120 million, representing 6% growth compared to Q2 2021.
- First Quarter 2021 and Recent Significant Corporate Highlights Commercial Performance ONPATTRO Achieved global net product revenues for the first quarter of 2021 of $102 million, representing 13% growth compared to Q4 2020.
- The Company reported preliminary global net product revenues for ONPATTRO and GIVLAARI for the fourth quarter and full year 2020 of approximately $112 million and $361 million, respectively. 2021 Financial Guidance Full year 2021 financial guidance consists of the following: Combined net product revenues for ONPATTRO, GIVLAARI, and OXLUMO $610 million - $660 million Net revenues from collaborations and royalties $150 million - $200 million GAAP R&D and SG&A expenses $1,335 million - $1,455 million Non-GAAP R&D and SG&A expenses* $1,175 million - $1,275 million *Excludes $160-$180 million of stock-based compensation from estimated GAAP R&D and SG&A expenses Use of Non-GAAP Financial Measures This press release contains non-GAAP financial measures, including expenses adjusted to exclude certain non-cash expenses and non-recurring gains outside the ordinary course of the Company's business.

- Our objective is to continue to execute successful product launches leveraging our positive experience with the launches of ONPATTRO, GIVLAARI and OXLUMO.

The neighbors of *onpattro* thus do more than identify the drug; they locate the specific register in which the company discusses it, and the boolean query recovered the exact sentences that register describes. The path from a single term, to its distributional neighbors, to the verbatim passages that produced them, is direct and auditable at every step.

Searching for *drones* in Mercury's semantic space returns *man_portable*, *sigint*, *isr*, *compact*, *broadband*, *phalanx*, *spy*, and *add_on*, whereas the same search in Kratos' space returns *missiles*, *jet*, *propulsion*, *munitions*, *loitering*, *supersonic*, *hypersonic*, *unmanned*, *combat*, and *engines*. The divergence reflects two different business models: in Mercury's filings *drones* sits among the sensor and signal-processing payloads that the company supplies, while in Kratos' it sits among complete airframes, propulsion systems, and munitions. The same word occupies two distinct regions of meaning, each faithful to how its issuer actually uses it, a distinction that follows directly from building each vector from its own corpus.

Retrieving the sentences that correspond to these neighbors in each space then provides a fully attributed basis for analysis. Because every term traces back to specific, well identified passages, the resulting extract can serve either as source material for a human analyst or as grounded input for a language model, one instructed to draw only on the supplied passages and to explicitly cite their identifiers. The word-level clustering and the sentence-level provenance together give the model exactly what it needs to remain faithful to the filings: the themes to organize its output around, and the verbatim, identifier-tagged text to quote from rather than paraphrase from memory.

## 6. Retrieval and corpus-relativity

The retrieval underlying this workflow can be examined directly and it is instructive to compare it against a conventional embedding approach. We do so with two queries, each run against a given sentence pool, contrasting the neighbors our method returns with those of a widely used sentence-embedding model.

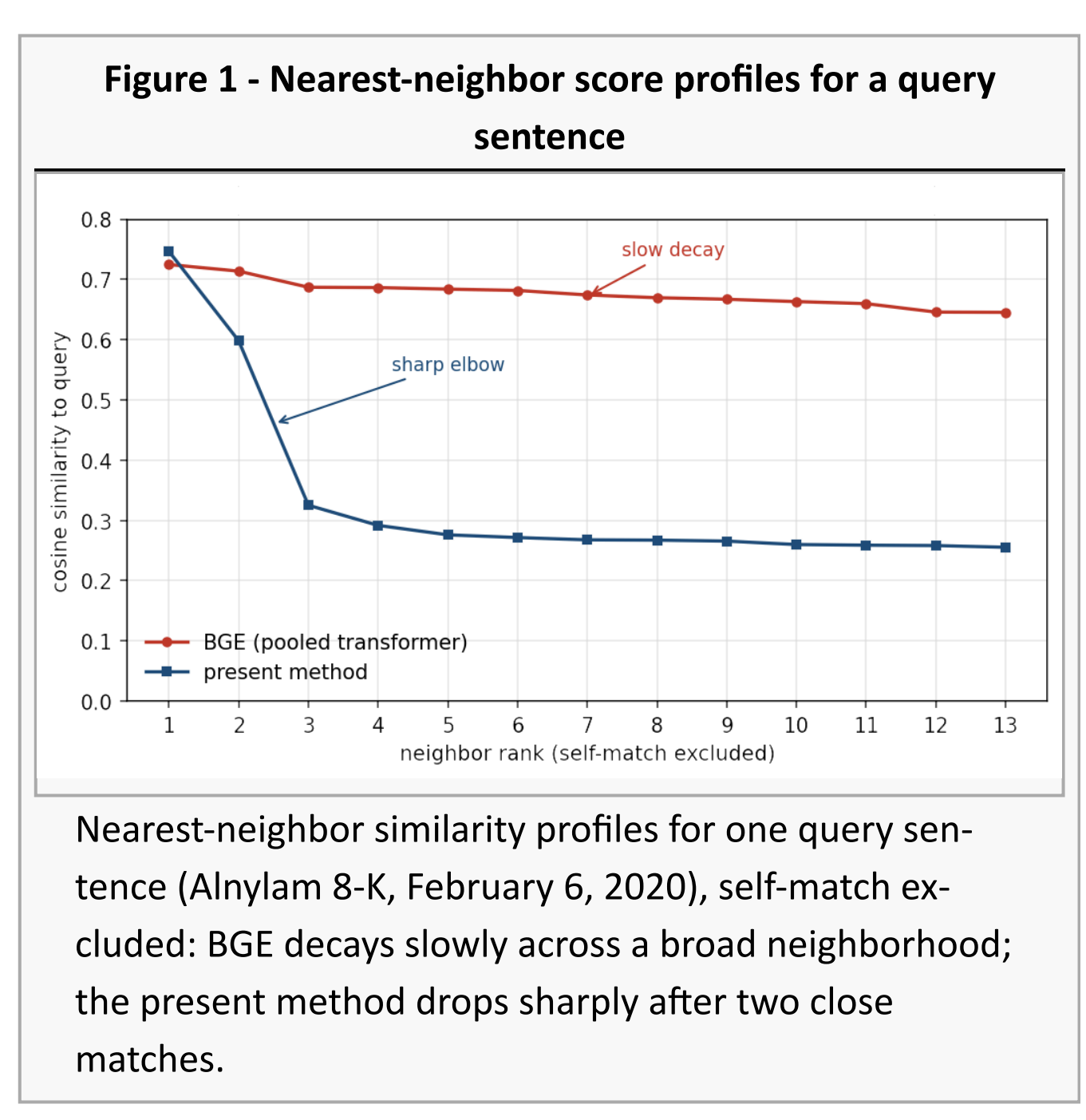

**Figure 1 - Nearest-neighbor score profiles for a query sentence**

Nearest-neighbor similarity profiles for one query sentence (Alnylam 8-K, February 6, 2020), self-match excluded: BGE decays slowly across a broad neighborhood; the present method drops sharply after two close matches.

In a first example (Fig. 1), both methods were queried with a sentence from the Alnylam February 6, 2020 8-K filing, ... *people afflicted with rare genetic, cardio-metabolic, hepatic infectious, and central nervous system (CNS)/ocular diseases...* , and searched the same 8-K's sentence pool. Each returned the query itself first, at simi-larity near 1.0, but then the two profiles differed characteristically. The pooled BGE transformer embedding (bge-small-en-v1.5) returned a broad topical neighborhood with slowly decaying scores from 0.72 through 0.65 across a dozen sentences ranging over trial enrollment, delivery technology, regulatory designations, and safety monitoring. The present method instead showed a sharp elbow: after the query, two sentences directly echo its vocabulary and score 0.75 and 0.60, but the similarity then falls abruptly to 0.33 and below, where a distinct secondary cluster concerning a single drug begins.

The two methods thus embody different, equally useful notions of 'nearest'. The transformer measures a diffuse semantic relatedness that groups the query with the filing's general clinical narrative, whereas the present method measures a vocabulary-anchored similarity that isolates the sentences genuinely close to the query and, through the elbow in its score profile, indicates where they end. Neither notion is inherently superior, but the sharper profile, with a clear cutoff, makes the present method better usable as a filter.

The text in this example came from an attachment to the Alnylam 8-K comprising 388.5 kB of HTML, reduced to 38.3 kB after tag removal and segmented into 168 sentences. Embedding and searching these sentences took 1.76 s with BGE on a GPU and 0.010 s with the present method on a single CPU core. While the precise ratio is hardware-dependent, the two-orders-of-magnitude difference in resource requirements reflects the far lower cost of a sequence of semantic-space lookups and vector additions compared with a transformer forward pass over the same text.

Polysemy is thus handled at the word level without any explicit disambiguation step. A word's semantic vector is not absolute: it is a function of the corpus that produced it, accumulating the contexts in which that corpus uses the word, so the same query returns different neighbors depending on which corpus's vectors are used. This is by design, as already noted in Section 2.2: a word's meaning is nothing more than its usage in the corpus at hand. *Digital* in a bank's filings and in a semiconductor maker's, or *cloud* as an accounting term and as a product line, are simply different vectors, each formed from its own contexts. The seed basis is shared across all corpora, so the vectors remain comparable, but the meaning each one carries is local to where it was formed.

In practice, of course, the most recent vectors should be used in most cases, but in order to demonstrate how the choice of different epoch's vectors changes focus, we used the following snippet as query: *... As part of this evaluation and approval process, the FAA had to make two distinct determinations: (i) whether the airplane met U.S. federal airworthiness standards; and (ii) what minimum level of pilot training would be required for a pilot to fly the airplane for a U.S.-based airline.* and used as a pool to be searched the first 200 abstracts returned by a Lucene search for "boeing" in our publications database. We varied only the Boeing vectors used to score them: first the semantic space accumulated to January 2021, the date of the source filing from which the query was extracted, then the space accumulated up to and including the March 2026 10-Q. Since query and pool are unchanged, any difference in the ranked neighbors is due solely to the choice of the vectors.

The two neighborhoods differ in exactly the way corpus-relativity predicts. Scored with the 2021 vectors, the query's neighbors (Fig. 2) are dominated by the 737 MAX crisis then defining Boeing's filings: certification and airworthiness news, shareholder litigation, and crash settlements rank highest. Scored with the 2026 vectors, the same sentence connects instead to Boeing's current landscape: the 777X charge, the twenty-year demand forecast, the Spirit AeroSystems acquisition, the Starliner contract, the defense-workers strike. The 2021 space reads the airworthiness determination through the lens of the crisis that then dominated the company's disclosures; the 2026 space reads it through the concerns of the present. The overall similarity scores are also lower under the 2026 vectors (the top neighbor falls from around 0.83 to 0.52), reflecting that against Boeing's larger and more varied 2026 vocabulary the specific 2021 query is less central.

**Figure 2 - The vector epoch as a lens: one query, one pool, two Boeing epochs**

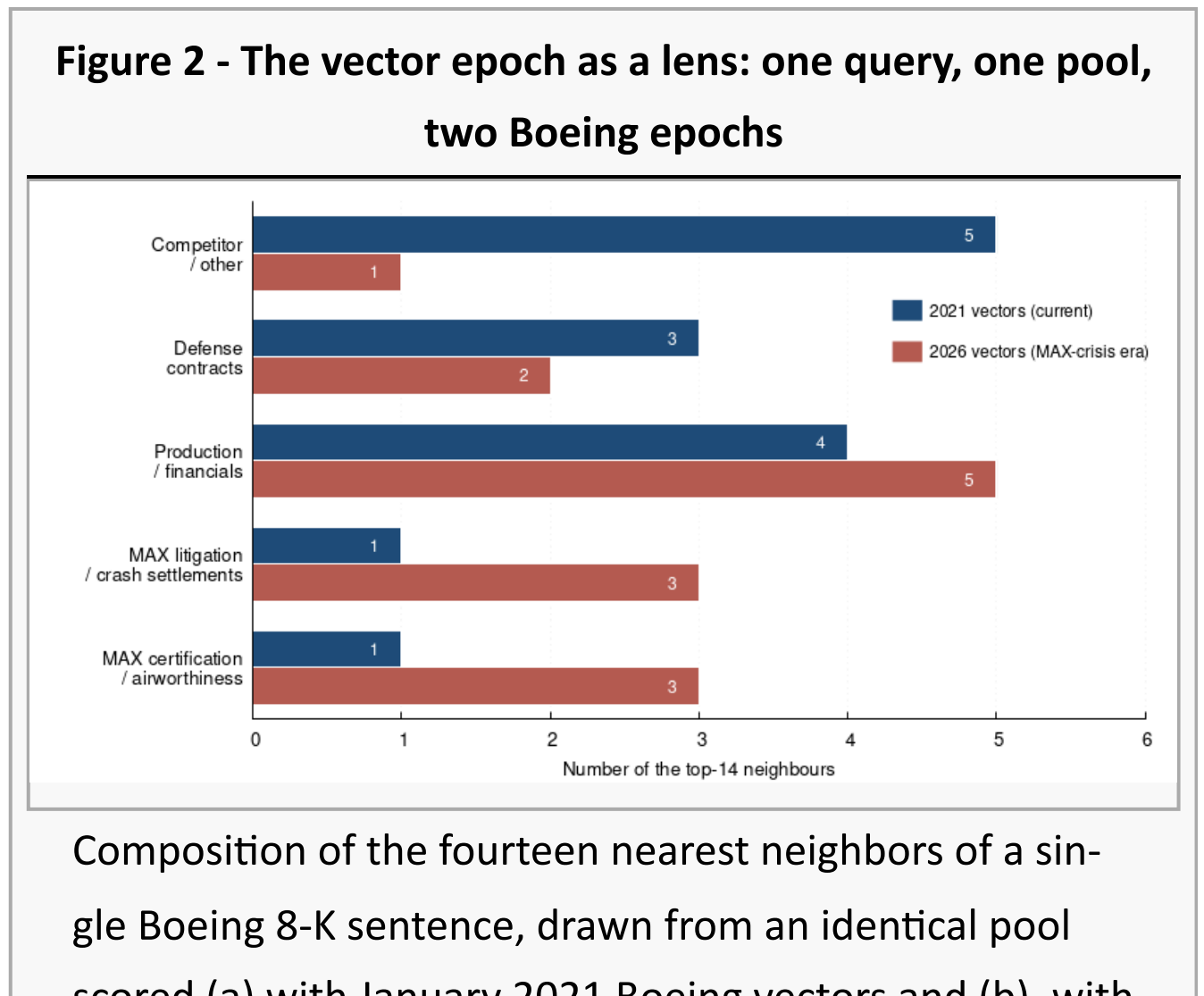

Composition of the fourteen nearest neighbors of a single Boeing 8-K sentence, drawn from an identical pool scored (a) with January 2021 Boeing vectors and (b), with March 2026 Boeing vectors. The 2021 vectors concentrate the neighbors on the 737 MAX crisis while the 2026 vectors spread them across Boeing's current landscape. Neighbor categories are assigned manually.

Neither result is more correct; each is faithful to the corpus that produced its vectors. What the comparison shows is that the epoch of the vectors is a controllable analytical parameter: the same query can be read through any point in an issuer's history, and because the vectors for any epoch are rebuilt from the raw filings in seconds, an analyst can, if needed for a retrospective study, sweep a query across successive years and watch its neighborhood evolve in support of historical analysis.

The same principle governs the choice of corpus more generally. To search within a corpus, one uses the vectors built from that corpus, and the epoch comparison above is the special case where the two corpora are the same issuer at different times. Searching one corpus with vectors imported from another issuer is a further, deliberate cross-space operation, useful precisely when one wants to ask how a concept native to one corpus falls within another, as in the projection of one issuer's vocabulary into a peer's space, but it is not the default and should be done knowingly.

This corpus-relativity is transparent to the user and imposes little burden on a search engine built on the method, because the semantic spaces form a nested family. Since the construction is purely additive, the space of a larger corpus already contains the contributions of every corpus within it, so a search can be served from the vectors of any space that includes its scope; the engine need maintain only spaces large enough to cover the corpora of interest, not one per possible query. In production this is simpler still: only the latest cumulative space of each issuer need be stored, as it incorporates every prior filing.

## 7. Conclusion

We have described a training-free, alignment-free method for representing and comparing corporate disclosures, in which each word is mapped by a deterministic hash to a fixed sparse seed vector and acquires meaning through the accumulation of its sentence contexts. Built on a single shared basis, the construction supports, at negligible cost and on ordinary hardware, a range of analyses we have illustrated on SEC filings: extracting an issuer's distinctive vocabulary, clustering a filing into its themes with verbatim supporting passages, comparing issuers and locating one company's concepts within another's disclosures, and tracking how an issuer's language evolves from one filing to the next.

The method is complementary to locally computed transformer or LLM-generated embeddings, which are in fact necessary for such tasks as sentiment scoring or the drafting of finished reports. Its performance for boolean retrieval is comparable to that of Lucene, except that it does not support fuzzy matching; on the other hand, it supports the retrieval of semantic neighbors, as demonstrated in Tables 1, 2, and 3.

The real strength of the method is that it is grounded in a single shared vector space, in which every object is represented on the same fixed basis. Heterogeneous sources can therefore be queried and compared together without conversion, alignment, or the information loss that pooling entails: words, sentences, filings, and whole corpora are all directly comparable and each source can be processed independently, yet any of them can be integrated at little cost, because they share the same basis.

In addition, nothing in the method being specific to SEC filings, the database is easily extended. Earnings calls, investor presentations, press releases, patent applications, medical and technical information, news feeds can be incorporated to enrich the retrieved information presented to the analyst or an LLM.

Two further properties follow from the construction. First, the method is deterministic and auditable: every score traces back to specific words and sentences in the source and a term absent from a corpus yields a clean null result rather than a plausible fabrication, an absolute guarantee generative models cannot offer. Second, when a language model is used at the end of the workflow, the extract it receives is already filtered, ordered, and grounded exclusively in the source text, with each passage uniquely identified, so the model can be constrained to quote rather than drawing on inferred knowledge. The determinism of the upstream extraction is what makes that grounding reliable, in a way no LLM prompt alone can guarantee.

The method makes no claim to answer an analyst's question on its own. Its role is upstream: to reduce a large and unstructured body of disclosure to a compact, searchable, and fully traceable database on which a human reader or a language model can operate more efficiently and more reliably. Because the extract is already filtered and grounded, a smaller in-house model or a less specialized analyst can be trusted with work that would otherwise demand a frontier model or a domain expert. The method faithfully reports the vocabulary and associations present in a filing, but it cannot on its own distinguish a substantive connection from an incidental one; that discrimination, as several of our examples showed, remains the reader's. Within these limits, what it offers is a fast, deterministic, and auditable foundation for corporate intelligence.

## Appendix

## A.1. Constructing the semantic space

### *A.1.1. Word weights*

Word weights in our construction are deliberately simple. Because each word occupies its own fixed direction in the seed basis rather than a position learned from the corpus, the usual TF-IDF weighting has no natural meaning here: there is no document-frequency term to discount, since a word's representation does not compete for space with others. We therefore derive weights directly from corpus frequency. Using the word counts in the Loughran-McDonald Master Dictionary [23], we set each word's weight to $w = \max(0, -\log(100\, f))$, where $f = n_w / n_{\text{the}}$ is the word's frequency normalized by the count of "the", scaled and stored as an integer. Common words thus receive low or zero weight and rare, informative words high weight, which is all the construction requires.

A curated list of boilerplate terms is treated apart, subtracting rather than adding weight, in three tiers by severity: the heaviest, applied to pure legal filler such as *hereunder* or *whereof*, is effectively exclusionary, suppressing any sentence built around such words; a middle tier, for forward-looking-statement and disclaimer vocabulary, strongly deprecates; and a light tier gently discounts generic filing terms.

### *A.1.2. From word to seed*

**Algorithm A.1** - Deterministic seed generation for a word (FNV-1a hash)

0 : Initialize the hash $h$ to the FNV offset basis
1 : **for each byte $c$ of the word, left to right**
2 : $h \leftarrow h \oplus c$
3 : $h \leftarrow h \times p$, where $p$ is the FNV prime (arithmetic modulo $2^{64}$)
4 : **return** the 64-bit hash $h$ as the word's seed

Each word is mapped to a 64-bit seed by the FNV-1a hash, a standard non-cryptographic hash chosen for being fast, deterministic, and well distributed. The seed serves two roles at once: it is the word's unique identifier in the database, and it is the value from which the word's sparse seed vector is generated. Because the mapping is purely deterministic, the same word always yields the same seed on any machine and at any time, which is what allows all documents and all epochs to share a single fixed basis without any coordination or storage of a vocabulary. The hash itself carries no meaning; it is simply a reproducible bridge from a string to a fixed point in the vector space, and meaning accrues only later, through the accumulation of context described below.

### *A.1.3. From seed to seed vector*

**Algorithm A.2** - Deterministic seed vector generation from a word's seed

0 : Initialize the vector $\mathbf{v}$ of dimension $D$ to all zeros
1 : Set the initial sign $s$ from the high-order bit of the seed
2 : **while fewer than $N$ non-zero entries have been placed**
3 : Advance the seed by one step of a deterministic bit-mixing map (xorshift)
4 : Derive a candidate index $j \leftarrow$ seed mod $D$
5 : **if** position $j$ is still zero
6 : Set $v_j \leftarrow s \cdot c$, where $c$ is the fixed non-zero magnitude
7 : Flip the sign, $s \leftarrow -s$, and count the entry as placed
8 : **return** the sparse seed vector $\mathbf{v}$

The seed vector is generated deterministically from the word's 64-bit seed, so that the same word always produces the same vector without any stored table. Starting from the all-zero vector, the seed is repeatedly passed through a fast bit-mixing step, each iteration yielding a candidate coordinate; the first $N$ distinct coordinates encountered are filled, alternating in sign, with a fixed magnitude. The result is a sparse vector with $N$ non-zero entries among $D$ dimensions, half positive and half negative, and the balance of signs keeps the vectors centered. Because $N \ll D$, two independently generated vectors rarely share a non-zero coordinate, which is precisely the quasi-orthogonality property that lets a modest dimension represent a large vocabulary with negligible interference. As with the seed, the construction is purely mechanical and carries no meaning of its own; it fixes each word at a reproducible point in the shared space, on which the accumulation of context then builds. In this paper, $D = 384$ and $N = 32$ have been used throughout. The fixed magnitude $c$ is set to $\frac{1}{\sqrt{32}}$ so as to generate a normalized vector.

### *A.1.4. From seed vector to semantic space*

This is the core accumulation function, where semantic vectors are built.

**Algorithm A.3** - Accumulating semantic vectors from a filing

0 : The semantic space is held in two stores: one mapping each seed to its word, one mapping each seed to its accumulated vector
1 : **if** this filing's identifier is already recorded, skip it; otherwise record it
2 : **for each sentence in the filing**
3 : Initialize the sentence vector $\mathbf{s}$ to zero
4 : **for each word in the sentence**
5 : Normalize the word; skip it if empty or a stop-word
6 : Compute its seed and generate its seed vector $\mathbf{e}$ from the fixed basis
7 : Record the word in the word store and add $w\,\mathbf{e}$ to $\mathbf{s}$, weighted by the word's weight $w$
8 : Remove duplicate words within the sentence
9 : **for each distinct word of the sentence**
10 : Add the sentence vector $\mathbf{s}$ into that word's accumulated vector in the vector store
11 : **return** the two updated stores

The construction proceeds sentence by sentence. Each sentence is first represented by the weighted sum of the seed vectors of its words, a frequent word contributing little and a rare one much; this is the sentence's own vector in the fixed basis. That sentence vector is then added into the accumulating semantic vector of every distinct word the sentence contains. A word's semantic vector is therefore the running sum of all the sentence contexts in which it has appeared, which is the concrete form taken by the distributional principle that a word is characterized by the company it keeps, with the full sentence as the context window. Both stores grow incrementally: a new filing adds its words and their contexts to what is already there, with no retraining and no recomputation of past filings, and the filing identifier is recorded so that the same document is never accumulated twice. The sentence vector added into each word's accumulated vector includes that word's own seed-vector contribution rather than excluding it; this self-inclusion is deliberate, retained after experimentation because it anchors each word's semantic vector to its own identity while still absorbing the surrounding context.

Each word's accumulated vector is updated once per sentence it appears in, regardless of how many times it occurs there, so that a word repeated within a single sentence is not thereby over-weighted; its running total is simply incremented by that sentence's vector.

### *A.1.5. Sentence vectors*

**Algorithm A.4** - Computing the vector of each sentence in a corpus

```
0 : Load the corpus semantic vectors from the hash store
1 : for each sentence in the corpus
2 :   Initialize the sentence vector s to zero
3 :   for each word in the sentence
4 :     Normalize and lowercase the word; skip it if empty or a stop-word
5 :     Compute its seed in order to look up its weight w; skip it if w ≤ 0
6 :     Retrieve the word's accumulated semantic vector from the store and normalize it to unit length
7 :     Accumulate: s ← s + w v̂
8 :   Normalize s to unit length and emit it as this sentence's vector
9 : return the set of sentence vectors
```

The vector of a sentence is the weighted sum of the unit-normalized semantic vectors of its words, drawn from the corpus store. Each word is first reduced to a canonical form and discarded if it is a stop-word or carries no weight; the remaining words contribute their accumulated semantic vector, scaled by a weight that is small for common terms and large for rare, informative ones, so that distinctive words dominate the sentence's representation. Normalizing each word vector before summation prevents a single high-magnitude term from dominating, and normalizing the final sum places every sentence vector on the unit sphere, so that sentences are directly comparable by inner product. Only the resulting sentence vectors are retained; the individual word vectors are read from the store as needed and not stored again.

## A.2. Further applications

### *A.2.1. Corporate fingerprints*

The profile (or fingerprint) of an issuer is the set of words most distinctive to its filings relative to a reference corpus. Each candidate word is scored from three factors: how often it occurs in the issuer's filings, its intrinsic weight (low for common terms, high for rare informative ones), and its frequency in the issuer corpus measured against the reference, using the count of the word "the" as a proxy for corpus size. The score retains only words that occur exclusively in this issuer's filings, that is, whose entire presence in the reference corpus comes from this issuer; these are the purest signature terms, present nowhere else, and they dominate the top of the ranking. Boilerplate, malformed tokens, and unweighted words are excluded, and the surviving words are ranked and truncated to form a compact fingerprint.

**Algorithm A.5** - Extracting an issuer's semantic profile

0 : Let $N_{\text{the}}^{\text{tick}}$ and $N_{\text{the}}^{\text{ref}}$ be the counts of *the* in the issuer and reference corpora
1 : **for each word $w$ in the reference store**
2 : Let $c_w^{\text{tick}}$ and $c_w^{\text{ref}}$ be its counts in the issuer and reference corpora
3 : Skip $w$ unless it occurs in both and $c_w^{\text{tick}} = c_w^{\text{ref}}$ (exclusive to this issuer)
4 : Skip $w$ if it is boilerplate, malformed, or carries no weight
5 : Score it: $s_w = c_w^{\text{tick}} \cdot \text{wht}_w \cdot \dfrac{N_{\text{the}}^{\text{ref}}}{N_{\text{the}}^{\text{tick}}}$
6 : Rank all scored words by $s_w$ and keep those above a threshold
7 : **return** the top-scoring words as the issuer's profile

### *A.2.2. Key-word extraction from a filing*

**Algorithm A.6** - Scoring the words a filing reinforces

0 : Given the candidate seeds present before the new filing
1 : **for each seed $w$**
2 : Retrieve its accumulated vector $\mathbf{v}_w^{\text{prev}}$ (before the filing) and $\mathbf{v}_w^{\text{cur}}$ (after)
3 : Score it by the *unnormalized* inner product $s_w = \langle \mathbf{v}_w^{\text{prev}}, \mathbf{v}_w^{\text{cur}} \rangle$
4 : **return** the seeds with their scores

For each word already present before a new filing, we measure how strongly that filing reinforced it by taking the inner product of the word's accumulated vector before the filing with its accumulated vector after. Crucially, this product is left unnormalized, since it is the magnitude of the reinforcement and not its direction that identifies the words a filing has amplified. Because the space is cumulative, a word untouched by the new filing scores near its prior magnitude, while a word the filing uses heavily gains a large aligned contribution and rises to the top; a word absent before the filing has no prior vector and cannot appear here, so this measure identifies the reinforcement of established vocabulary rather than the introduction of new terms, the latter being caught by the frequency-based measure previously described.

### *A.2.3. Named-entity extraction*

To identify salient multiword named entities and other recurring expressions in SEC filings, we apply the discounted collocation score of Mikolov et al. [6] to adjacent token pairs. For two consecutive tokens $w_0$ and $w_1$, the score is:

$$\text{score}(w_0, w_1) = \frac{\text{count}(w_0 w_1) - \delta}{\text{count}(w_0)\,\text{count}(w_1)} \tag{A.1}$$

Here, $\text{count}(w_0 w_1)$ is the number of occurrences of the ordered adjacent pair in the filing corpus, while $\text{count}(w_0)$ and $\text{count}(w_1)$ are the corresponding unigram counts. Pairs whose score exceeds a specified threshold are concatenated with an underscore.

The numerator includes a discount parameter, $\delta$, to reduce spurious high scores for rare word pairs. Without this discount, a pair observed only once can appear disproportionately distinctive when both individual words are rare. Substracting $\delta$ creates an effective minimum-recurrence requirement: a pair must occur more than $\delta$ times before it can obtain a positive score.

In SEC filings, the resulting multiword tokens can capture named entities, product names, regulatory concepts, financial instruments, and recurring legal or accounting constructs.

### *A.2.4. Filing complexity*

**Algorithm A.7** - Effective rank of an issuer's fingerprint

0 : Load the issuer's fingerprint: its top $N$ distinctive words with weights $w_i$
1 : Build the sentence vector $\mathbf{v}_i$ of each fingerprint word
2 : Normalize the weights so that $\sum_i w_i = 1$
3 : Form the weighted covariance $C$ of the vectors $\{\mathbf{v}_i\}$ about their weighted mean
4 : **return** the participation ratio $\left(\sum_k \lambda_k\right)^2 / \sum_k \lambda_k^2$ of the eigenvalues $\lambda_k$ of $C$

In practice the eigenvalues are never computed explicitly: both sums are obtained from the traces of $C$ and $C^2$, the latter reducing to the pairwise dot products $(\mathbf{v}_i \cdot \mathbf{v}_j)^2$, so the effective rank costs only $O(N^2)$ operations over the $N$ fingerprint words.

The effective rank of an issuer's fingerprint [24] ranges from $1$, when all the word vectors point in a single direction, up to the number of vectors, when they are mutually orthogonal, and so measures the number of effectively independent directions the fingerprint occupies: a focused, single-line issuer yields a low value while a thematically diverse one yields a high value. This index of filing complexity is derived entirely from the text, giving a simple, deterministic, filing-level scalar independent of text length. We report it as an index in its own right; whether it complements any established notion of firm or disclosure complexity [22] is a question we leave open.

The extremes (next page) are easy to interpret: the lowest values correspond to focused, single-line issuers (a logistics REIT, a regulated utility, an oilfield-services firm, a single-platform biotech), the highest to issuers whose distinctive vocabulary spans many themes. Clearly, a globally complex industrial company could have a semantically narrow filing fingerprint if its high-distinctiveness terms concentrate on one proprietary technology or operating theme. Conversely, a smaller firm could have a broad semantic fingerprint because it has contracts, litigation, acquisitions, platform terminology, product lines, and government relationships.

Computed over 30 issuers from our corpus, with each fingerprint truncated to its top 50 words, the index yields:

| Table A.1 - Filing complexity of selected corporations | | | | | |
|---|---|---|---|---|---|
| **1. Palantir** | *25.8* | **2. Kratos** | *22.3* | **3. Amazon** | *20.8* |
| **4. Walmart** | *19.6* | **5. Mercury** | *17.5* | **6. Johnson & J.** | *17.1* |
| **7. Vertex** | *16.6* | **8. Apple** | *16.4* | **9. Moderna** | *16.1* |
| **10. Delta** | *14.9* | **11. Pfizer** | *13.8* | **12. Merck** | *13.8* |
| **13. AeroVironment** | *13.7* | **14. Intel** | *13.7* | **15. Target** | *13.1* |
| **16. Archer-Daniels-M.** | *12.8* | **17. Broadcom** | *12.4* | **18. Chevron** | *12.3* |
| **19. Bristol-Myers** | *11.7* | **20. NVIDIA** | *11.6* | **21. Incyte** | *10.3* |
| **22. Bunge** | *10.2* | **23. NIKE** | *10.0* | **24. Amgen** | *9.9* |
| **25. Eli** | *9.7* | **26. AMD** | *9.4* | **27. Caterpillar** | *7.7* |
| **28. Alnylam** | *4.7* | **29. Schlumberger** | *4.3* | **30. NextEra** | *4.0* |